\documentclass[runningheads]{llncs}
\usepackage{graphicx}

\usepackage{tikz}
\usepackage{comment}
\usepackage{amsmath,amssymb} 
\usepackage{color}
\usepackage[ruled]{algorithm2e}
\usepackage{multirow}
\usepackage{subfigure}
\usepackage{bm}
\usepackage{wrapfig}
\usepackage{marvosym}
\allowdisplaybreaks[4]

\usepackage[accsupp]{axessibility}  

\begin{document}
\pagestyle{headings}
\mainmatter
\def\ECCVSubNumber{6441}  

\title{Black-Box Dissector: Towards Erasing-based Hard-Label Model Stealing Attack} 

\titlerunning{Black-Box Dissector}
%
\author{Yixu Wang\inst{1} \and
Jie Li\inst{1} \and
Hong Liu\inst{2} \and
Yan Wang\inst{3} \and
Yongjian Wu\inst{4} \and 
Feiyue Huang\inst{4} \and
Rongrong Ji\inst{1,5}$^\textrm{\Letter}$
}
\authorrunning{Wang et al.}
%
\institute{Media Analytics and Computing Lab, School of Informatics, Xiamen University \and
National Institute of Informatics \and
Pinterest \and
Youtu Lab, Tencent Technology (Shanghai) Co.,Ltd \and
Institute of Artificial Intelligence, Xiamen University \\
\email{yxwang79@gmail.com,}
\email{lijie.32@outlook.com,}
\email{hliu@nii.ac.jp,}
\email{yanw@pinterest.com,}
\email{\{littlekenwu,garyhuang\}@tencent.com,}
\email{rrji@xmu.edu.cn}
}
\maketitle

\begin{abstract}
Previous studies have verified that the functionality of black-box models can be stolen with full probability outputs.
However, under the more practical hard-label setting, we observe that existing methods suffer from catastrophic performance degradation.
We argue this is due to the lack of rich information in the probability prediction and the overfitting caused by hard labels.
To this end, we propose a novel hard-label model stealing method termed \emph{black-box dissector}, which consists of two erasing-based modules.
One is a CAM-driven erasing strategy that is designed to increase the information capacity hidden in hard labels from the victim model. The other is a random-erasing-based self-knowledge distillation module that utilizes soft labels from the substitute model to mitigate overfitting.
Extensive experiments on four widely-used datasets consistently demonstrate that our method outperforms state-of-the-art methods, with an improvement of at most $8.27\%$.
We also validate the effectiveness and practical potential of our method on real-world APIs and defense methods.
Furthermore, our method promotes other related tasks, \emph{i.e.},  transfer adversarial attacks.
\end{abstract}

\section{Introduction}

\begin{figure}[t]
    \centering
    \includegraphics[width=\textwidth]{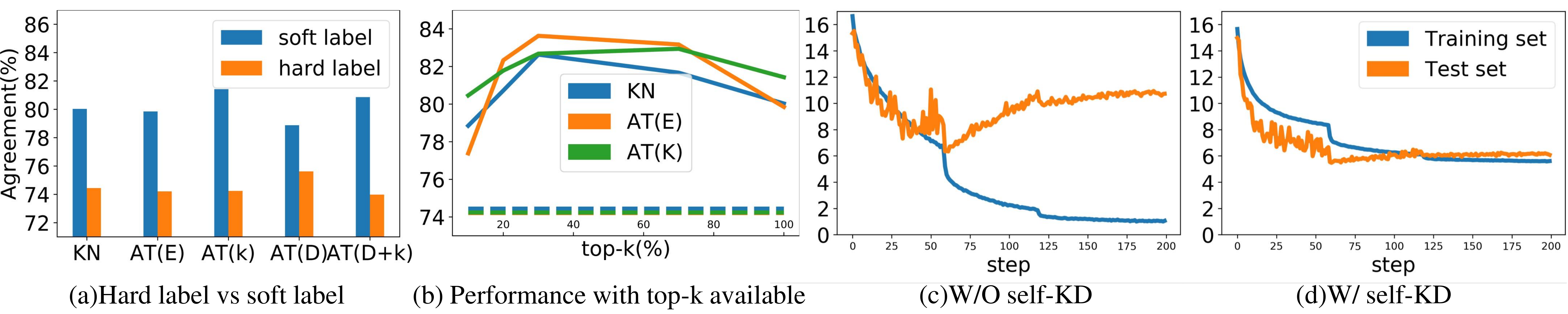}
    \caption{(a) The test accuracies of previous methods with hard labels are much lower than the ones with soft labels. (KN: KnockoffNets, `AT': ActiveThief, `E': entropy, `K': k-Center, `D': DFAL) (b) The performance decreases as the number of available classes decreases (dotted line : hard-label setting). (c) $\&$ (d) Loss curves for training/test set during model training without and with self-KD. All results are on the CIFAR10 dataset.}
    \label{fig1}
\end{figure}

Machine learning models deployed on the cloud can serve users through the application program interfaces (APIs) to improve productivity.
Since developing these cloud models is a product of intensive labor and monetary effort, these models are valuable intellectual property and AI companies try to keep them private~\cite{kariyappa2020defending,orekondy2019prediction,maini2021dataset,wang2020information}.
%
However, the exposure of the model’s predictions represents a significant risk as an adversary can leverage this information to steal the model’s functionality, \emph{a.k.a.} model stealing attack~\cite{papernot2017practical,orekondy19knockoff,pal2020activethief,ijcai2021-336,yu2020cloudleak}.
With such an attack, adversaries are able to not only use the stolen model to make a profit, but also mount further adversarial attacks~\cite{zhou2020dast,yang2020learning}.
Besides, the model stealing attacks is a kind of black-box knowledge distillation which is a hot research topic.
Studying various mechanisms of model stealing attack is of great interest both to AI companies and researchers.

Previous methods~\cite{orekondy19knockoff,zhou2020dast,pal2020activethief,ijcai2021-336} mainly assume the complete probability predictions of the victim model available,
while the real-world APIs usually only return partial probability values (top-$k$ predictions) or even the top-1 prediction (\emph{i.e.}, hard label).
In this paper, we focus on the more challenging and realistic scenario, \emph{i.e.}, the victim model only outputs the hard labels.
However, under this setting, existing methods suffer from a significant performance degradation, even by 30.50\% (as shown in the Fig.\,\ref{fig1} (a) and the appendix Tab.\,\uppercase\expandafter{\romannumeral1}). 
%

To investigate the reason for the degradation, we evaluate the performance of attack methods with different numbers of prediction probability categories available and hard labels as in Fig.\,\ref{fig1} (b).
With the observation that the performance degrades when the top-$k$ information missing,
we conclude that the top-$k$ predictions are informative as it indicates the similarity of different categories or multiple objects in the picture,
and previous attack methods suffer from such information obscured by the top-$1$ prediction under the hard-label setting.
It motivates us to re-mine this information by eliminating the top-$1$ prediction.
Particularly, we design \textit{a novel CAM-based erasing method},
which erases the important area on the pictures based on the substitute model's top-1 class activation maps (CAM)~\cite{selvaraju2017grad,zhou_learning_2016} and queries the victim model for a new prediction. 
%
Note that we can dig out other class information in this sample if the new prediction changes. 
Otherwise, it proves that the substitute model pays attention to the wrong area. 
Then we can align the attention of the substitute and the victim model by learning clean samples and the corresponding erased samples simultaneously. 

Besides, previous works on the self-Knowledge Distillation (self-KD)~\cite{kim2020self}, calibration~\cite{guo2017calibration}, and noisy label~\cite{zhang2016understanding} have pointed out 
the hard and noisy labels 
will introduce overfitting and miscalibration. 
More specifically, the attack algorithms cannot access the training data, and thus can only use the synthetic data or other datasets as a substitute, which is noisy. 
Therefore, the hard-label setting will suffer from overfitting, which leads to worse performance, and we verify it by plotting the loss curves in Fig.\,\ref{fig1} (c). 
To mitigate this problem, we introduce \textit{a simple self-knowledge distillation module with random erasing (RE)} to utilize soft labels for generalization. 
Particularly, we randomly erase one sample a certain number of times, query the substitute model for soft-label outputs, and take the average value of these outputs as the pseudo-label. 
After that, we use both hard labels from the victim model and pseudo labels from the previous substitute model to train a new substitute model. 
Therefore, we can also consider the ensemble of the two models as the teacher in knowledge distillation. 
As in Fig.\,\ref{fig1} (d), such a module helps generalization and better performance.


\begin{figure}[t]
    \centering
    \includegraphics[width=\textwidth]{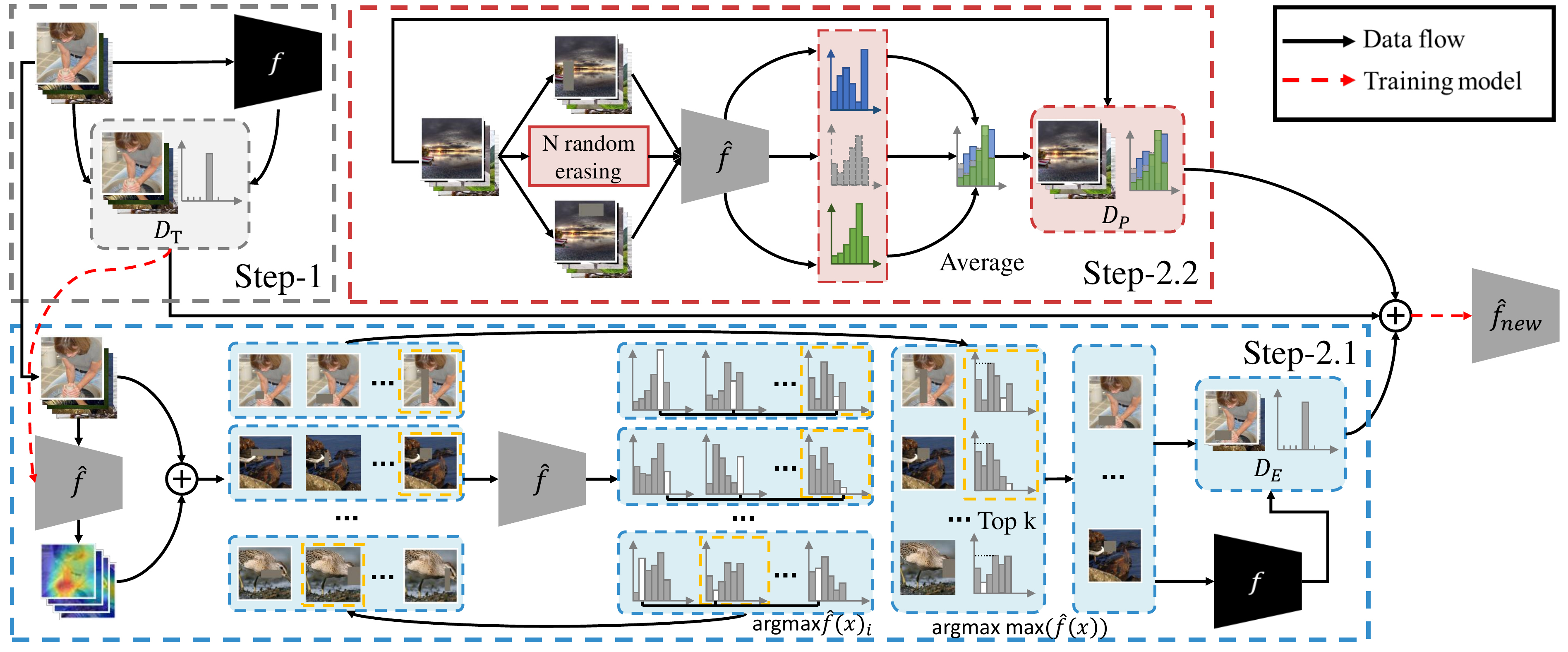}
    \caption{Details of our proposed black-box dissector with a CAM-driven erasing strategy (step 2.1) and a RE-based self-KD module (step 2.2). 
    In step 2.1, the images in transfer set $D_T$ are erased according to the Grad-CAM, and we selected the erased images with the largest difference from the original images according to the substitute model's outputs. 
    In step 2.2, we randomly erase the unlabeled image $N$ times, and then average the outputs of the $N$ erased images by the substitute model as the pseudo-label.}
    \label{fig:overview}
\end{figure}

In summary, we propose a novel model stealing framework termed \emph{black-box dissector}, which includes a CAM-driven erasing strategy and a RE-based self-KD module.
Our method is orthogonal to previous approaches~\cite{orekondy19knockoff,pal2020activethief} and can be integrated with them.
The experiments on four widely-used datasets demonstrate our method achieves $43.04-90.57\%$ test accuracy ($47.60-91.37\%$ agreement) to the victim model, which is at most $8.27\%$ higher than the state of the art method.
We also proved that our method can defeat popular defense methods and is effective for real-world APIs like services provided by Amazon Web Services (AWS).
Furthermore, our method promotes downstream tasks, \emph{i.e.}, transfer adversarial attack, with $4.91\%-16.20\%$ improvement.

\section{Background and Notions}
\textbf{Model stealing attack} aims to find a substitute model $\hat{f} \colon [0,1]^d \mapsto \mathbb{R}^{N}$ that performs as similarly as possible to the victim model $f \colon [0,1]^{d} \mapsto \mathbb{R}^{N}$ (with only outputs accessed). 
\cite{papernot2017practical} first observed that online models could be stolen through multiple queries. 
%
After that, due to the practical threat to real-world APIs, several studies paid attention to this problem and proposed many attack algorithms.

These algorithms consist of two stages: 1) constructing a transfer dataset $D_T$ (step 1 in Fig.\,\ref{fig:overview}) and 2) training a substitute model. 
The transfer dataset is constructed based on data synthesis or data selection and then feed into the victim model for labels. 
Methods based on data synthesis~\cite{zhou2020dast,kariyappa2020maze,NEURIPS2020_e8d66338} adopt the GAN-based models to generate a virtual dataset. 
And the substitute model and the GAN model are trained alternatively on this virtual dataset by querying the victim model iteratively. 
The data selection methods prepare an attack dataset as the data pool, and then sample the most informative data via machine learning algorithms, \emph{e.g.}, reinforcement learning~\cite{orekondy19knockoff} or active learning strategy~\cite{pal2020activethief}, uncertainty-based strategy~\cite{lewis1994sequential}, k-Center strategy~\cite{sener2018active}, and DFAL strategy~\cite{ducoffe2018adversarial}. 
Considering that querying the victim model will be costly,
the attacker usually sets a budget on the number of the queries, so the size of the transfer dataset should be limited as well. 
%
Previous methods assume the victim model returns a complete probability prediction $f(x)$, which is less practical.

In this paper, we focus on a more practical scenario that is about hard-label $\phi(f(x))$ setting, where $\phi$ is the truncation function used to truncate the information contained in the victim's output and return the corresponding one-hot vector:
\begin{equation}
    {\phi(f(x))}_{i}:=
    \begin{cases}
    1 & \text{if } i = \mathop{\arg\max}_{n}{f(x)}_{n} \,; \\
    0 & \text{otherwise} \,.                               \\
    \end{cases}
\end{equation}
With the transfer dataset, the substitute model is optimized by minimizing a loss function $\mathcal{L}$ (\emph{e.g.,} cross-entropy loss function):
\begin{equation}
\begin{cases}
    \mathbb{E}_{x \sim{\mathcal{D}_{T}}}\big[\mathcal{L}\big(f(x),\hat{f}(x)\big)\big], &\text{for soft labels}; \\
    \mathbb{E}_{x \sim{\mathcal{D}_{T}}}\big[\mathcal{L}\big(\phi(f(x)),\hat{f}(x)\big)\big], &\text{for hard labels}. \\
\end{cases}
\end{equation}
%


\textbf{Knowledge distillation} (KD) has been widely studied in machine learning~\cite{hinton2015distilling,anil2018large,furlanello2018born}, which transfers the knowledge from a teacher model to a student model. 
Model stealing attacks can be regarded as a black-box KD problem where the victim model is the \textit{teacher} with only outputs accessible and the substitute model is the \textit{student}. 
The main reason for the success of KD is the \textit{valuable information that defines a rich similarity structure over the data} in the probability prediction~\cite{hinton2015distilling}.
However, for the hard-label setting discussed in this paper, this valuable information is lost. 
And the main difference between self-KD and regular-KD is that the latter utilizes knowledge from a larger and better teacher model, while the former uses the model self as the teacher.
Self-KD has been shown to help improve the model's generalization ability~\cite{kim2020self}. 
Inspired by self-KD, our method tries to dig out the hidden information in the data and models, and then transfers more knowledge to the substitute model.

\textbf{The erasing-based method}, \emph{e.g.}, random erasing (RE)~\cite{zhong2020random,devries2017cutout}, is currently one of the widely used data augmentation methods, which generates training images with various levels of occlusion, thereby reducing the risk of over-fitting and improving the robustness of the model. Our work is inspired by RE and designs a prior-driven erasing operation, which erases the area corresponding to the hard label to re-mine missing information.

\section{Method}
The overview of black-box dissector is shown in Fig.\,\ref{fig:overview}.
In addition to the conventional process (\emph{i.e.}, the transfer dataset $D_T$ constructing in step 1 and the substitute model training in the right), 
we introduce two key modules: a CAM-driven erasing strategy (step 2.1) and a RE-based self-KD module (step 2.2).

\begin{wrapfigure}[20]{R}{0.5\textwidth}
\centering
\includegraphics[width=0.5\textwidth]{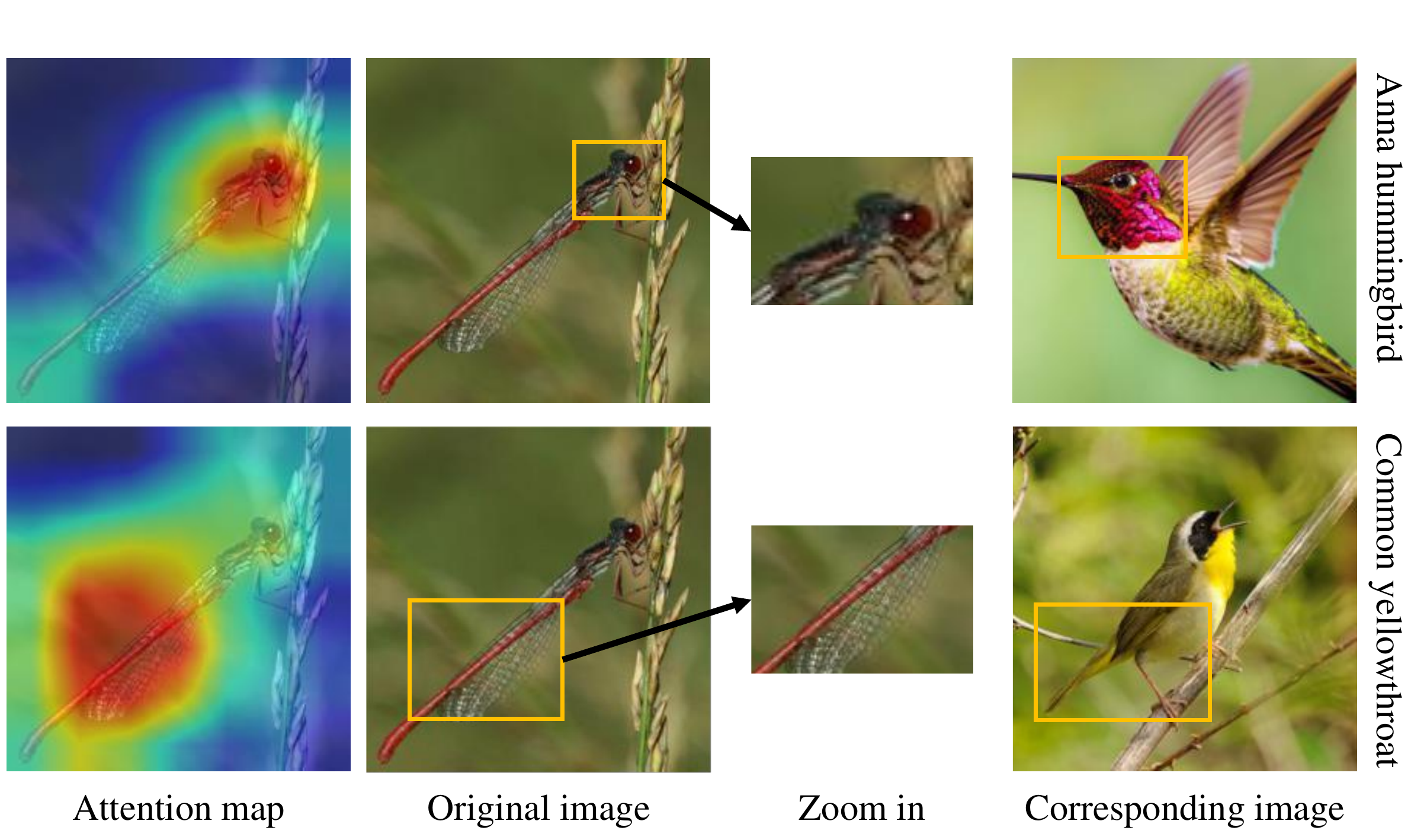}
\caption{An example from the ILSVRC-2012 dataset and its attention map corresponding to two most likely class ``Anna humming bird" and ``Common yellow throat" on the CUBS200 trained model. 
The attention areas share similar visual apparent with images of ``Anna humming bird" and ``Common yellow throat", respectively.}
\label{fig2}
\end{wrapfigure}

\subsection{A CAM-driven erasing strategy}
Since the lack of class similarity information degrades the performance of previous methods under the hard-label setting, we try to re-dig out such hidden information.
%
Taking an example from the ILSVRC-2012 dataset for illustration as in Fig.\,\ref{fig2}. 
Querying the CUBS200 trained victim model with this image, we get two classes with the highest confidence score:  ``Anna hummingbird" (0.1364) and ``Common yellowthroat" (0.1165), and show their corresponding attention map in the first column of Fig.\,\ref{fig2}. 
%
It is easy to conclude that two different attention regions response for different classes according to the attention map. 
When training the substitute model with the hard label ``Anna hummingbird" and without the class similarity information, the model can not learn from the area related to the ``Common yellowthroat" class, which means this area is wasted. 
To re-dig out the information about the ``Common yellowthroat" class, we need to erase the impact of the ``Anna hummingbird" class.

To this end, a natural idea is to erase the response area corresponding to the hard label. 
Since the victim model is a black-box model, we use the substitute model to approximately calculate the attention map instead. 
If the attention map calculated by the substitute model is inaccurate and the victim model's prediction on the erased image does not change, although we cannot obtain the class information, we can align the attention map of two models by letting the substitute model learn the original image and the erased one simultaneously. 
The attention maps can be used as sources of additional supervision signal in distillation: encouraging a model's attention map to be similar to that of another model also leads to the models having similar predictions~\cite{zagoruyko2016paying}. 
To get the attention map, we utilize the Grad-CAM~\cite{selvaraju2017grad} in this paper.
With the input image $x \in [0,1]^{d}$ and the trained DNN $\mathcal{F} \colon [0,1]^{d} \mapsto \mathbb{R}^{N}$, we let $\alpha_{k}^{c}$ denote the weight of class $c$ corresponding to the $k$-th feature map, 
and calculate it as $\alpha_{k}^{c} = \frac{1}{Z}\sum_{i}\sum_{j} \frac{\partial{\mathcal{F}(x)}^{c}} {\partial{A}^{k}_{ij}}, $ where $Z$ is the number of pixels in the feature map, ${\mathcal{F}(x)}^{c}$ is the score of class $c$ and ${A}^{k}_{ij}$ is the value of pixel at $(i,j)$ in the $k$-th feature map. After obtaining the weights corresponding to all feature maps, the final attention map can be obtained as $S^{c}_{\mathrm{Grad-CAM}} = \mathrm{ReLU} (\sum_{k} \alpha_{k}^{c} {A}^{k})$ via the weighted summation.

\begin{algorithm}[t]
\caption{Prior-driven Erasing $\psi(I, P)$}
\label{alg2}
\LinesNumbered
\KwIn{Input image $I$, prior probability $P$, area of image $S$, erasing area ratio range $s_l$ and $s_h$, erasing aspect ratio range $r_1$ and $r_2$.}
\KwOut{Erased image ${I}^{\prime}$.}
$S_e \sim \mathrm{Uniform}(s_l, s_h) \times S$, 
$ r_e \sim \mathrm{Uniform}(r_1, r_2)$ \\
$H_e \leftarrow \sqrt{S_e \times r_e}/2, W_e \leftarrow \sqrt{\frac{S_e}{r_e}}/2$ \\
$x_e, y_e$ sampled randomly according to $P$ \\
$I_e \leftarrow (x_e - W_e, y_e - H_e, x_e + W_e, y_e + H_e)$ \\
$I(I_e) \sim \mathrm{Uniform}(0, 255)$ \\
${I}^{\prime} \leftarrow I$ \\
\end{algorithm}

To erase the corresponding area, inspired by~\cite{zhong2020random}, we define a prior-driven erasing operation as $\psi(I, P)$, shown in Alg.\,\ref{alg2}, which randomly erases a rectangle region in the image $I$ with random values while the central position of the rectangle region is randomly selected following the prior probability $P$. 
The prior probability $P$ is of the same size as the input image and is used to determine the probability of different pixels being erased. 
%
Here, we use the attention map from Grad-CAM as the prior.
Let $x \in [0,1]^{d}$ denote the input image from the transfer set and $S^{\mathop{\arg \max}\hat{f}(x)}_{\mathrm{Grad-CAM}}(x, \hat{f})$ denote the attention map of the substitute model $\hat{f}$. 
This CAM-driven erasing operation can be represented:
\begin{equation}
    \psi\left(x, S^{\mathop{\arg\max} \hat{f}(x)}_{\mathrm{Grad-CAM}}(x, \hat{f})\right).
\end{equation}
We abbreviate it as $\psi(x, S(x,\hat{f}))$. 
To alleviate the impact of inaccurate Grad-CAM caused by the difference between the substitute model and the victim one,
for each image, we perform this operation $N$ times ($\psi_i$ means the $i$-th erasing) and select the one with the largest difference from the original label. Such a data augment operation helps the erasing process to be more robust.

We use the cross-entropy to calculate the difference between the new label and the original label, and we want to select the sample with the biggest difference. 
Formally, we define $\Pi(x)$ as the function to select the most different variation of image $x$:

\begin{equation}
\label{eq9}
\begin{aligned}
       & \Pi(x) := \psi_{k}(x,S(x,\hat{f})), \\
       \text{where } k := & \mathop{\arg\max}_{i \in [N]} - \sum_{j} {\phi\left(f\left(x\right)\right)}_{j} \cdot \log \left({\hat{f}\big(\psi_{i}(x,S(x,\hat{f}))\big)}_{j}\right) \\
       = & \mathop{\arg\max}_{i \in [N]} - \log \left({\hat{f}\big(\psi_{i}(x, S(x,\hat{f}))\big)}_{\mathop{\arg\max} \phi\big(f(x)\big)} \right) \\
       = & \mathop{\arg\min}_{i \in [N]} {\hat{f}\left(\psi_{i}(x, S(x,\hat{f}))\right)}_{\mathop{\arg\max} {\phi\big(f(x)\big)}} .
\end{aligned}
\end{equation}

Due to the limitation of the number of queries, we cannot query the victim model for each erased image to obtain a new label. 
We continuously choose the erased image with the highest substitute's confidence until reaching the budget. 
To measure the confidence of the model, we adopt the Maximum Softmax Probability (MSP) for its simplicity:
\begin{equation}
\label{eq10}
 \mathop{\arg\max}_{x\sim{\mathcal{D}_{T}}} {\hat{f}\left(\Pi\left(x\right)\right)}_{\mathop{\arg\max} {\hat{f}\left(\Pi\left(x\right)\right)}},
\end{equation}
where $D_{T}$ is the transfer set. The erased images selected in this way are most likely to change the prediction class. Then, we query the victim model to get these erased images' labels and construct an erased sample set $D_E$. 
Note that the substitute model is trained with $D_{T}$ to fit the victim model, so it makes sense to use the substitute model to approximately calculate the Grad-CAM. 
For each sample, if the approximate calculated Grad-CAM is accurate, it means we have erased the correct area, and we can obtain new class information after querying the victim model. 
If the Grad-CAM is inaccurate, it means the substitute model has paid attention to the wrong area, and we can align the attention map of two models by letting the substitute model learn the original image and the erased one simultaneously. 
Therefore, regardless of the accuracy of the area we erased, the erased sample can provide information to help the substitute model better approximate the victim model. 
We show the effect of our method to align the attention map in Fig.\,\ref{fig:cam}.

\begin{algorithm}[t]
\caption{Black-box Dissector}
\label{alg1}
\LinesNumbered
\KwIn{Unlabeled pool $D_U$, victim model $f$, maximum number of queries $Q$.}
\KwOut{Substitute model $\hat{f}$.}
Initialize $q \leftarrow 0, D_T \leftarrow \varnothing, D_E \leftarrow \varnothing$  \\
\While{$q<Q$}{
\textbf{// Step 1} \\
Select samples from $D_U$ according to budget and query $f$ to updata $D_T$ \\
$q = q + \text{budget}$ \\
$\mathcal{L}=\sum_{x \in D_T} \mathcal{L}^{\prime}\big(\phi(f(x)),\hat{f}(x)\big)$ \\
$\hat{f} \leftarrow update(\hat{f}, \mathcal{L})$ \\
\textbf{// A CAM-driven erasing strategy (step 2.1)} \\
Erase samples in $D_T$ according to Eq.\,\ref{eq9}  \\
Choose samples from erased samples according to Eq.\,\ref{eq10} and budget \\
Query $f$ to get labels and update $D_E$ \\
$\mathcal{L}=\sum_{x \in D_T \cup D_E} \mathcal{L}^{\prime}\big(\phi(f(x)),\hat{f}(x)\big)$ \\
$\hat{f} \leftarrow update(\hat{f}, \mathcal{L})$ \\
$q = q + \text{budget}$ (Check if $q<Q$) \\
\textbf{// A random-erasing-based self-KD (step 2.2)} \\
Select samples from $D_U$ \\
Get pseudo-labels according to Eq.\,\ref{eq11} and construct a pseudo-label set $D_P$ \\
$\mathcal{L}= \sum_{x \in D_T \cup D_E} \mathcal{L}^{\prime}\big(\phi(f(x)),\hat{f}(x)\big) + \sum_{x \in D_P} \mathcal{L}^{\prime}\big(y_p(x,\hat{f}) , \hat{f}(x)\big)$ \\
$\hat{f} \leftarrow update(\hat{f}, \mathcal{L})$ \\
}
\end{algorithm}

\subsection{A random-erasing-based self-KD module}
We also find that in training with limited hard-label OOD samples, the substitute model is likely to overfit the training set, which damages its generalization ability~\cite{kim2020self,zhang2016understanding}. 
Therefore, based on the above erasing operation, we further design a simple RE-based self-KD method to improve the generalization ability of the substitute model.

Formally, let $x \in [0,1]^{d}$ denote the unlabeled input image. We perform the erasing operation with a uniform prior $U$ on it $N$ times, and then average the substitute's outputs on these erased images as the original image's pseudo-label:
\begin{equation}
\label{eq11}
    y_{p}(x,\hat{f}) = \frac{1}{N} \sum_{i=1}^N \hat{f}\big(\psi_{i}(x, U)\big).
\end{equation}

This is a type of consistency regularization, which enforces the model to have the same predictions for the perturbed images and enhances the generalization ability. 
With Eq.\ref{eq11}, we construct a new soft pseudo label set $D_P=\{\big(x, y_p(x,\hat{f})\big),\dots\}$.

With the transfer set $D_T$, the erased sample set $D_E$, and the pseudo-label set $D_P$, we train new substitute model using the ensemble of the victim model and the previous substitute model as the teacher. 
Our final objective function is:
\begin{equation}
    \min \mathcal{L}= \min \big[\sum_{x \in D_T \cup D_E} \mathcal{L}^{\prime}\big(\phi(f(x)),\hat{f}(x)\big) \\ + \sum_{x \in D_P}  \mathcal{L}^{\prime}\big(y_p(x,\hat{f}) , \hat{f}(x)\big) \big]. \\
\end{equation}
where $\mathcal{L}^{\prime}$ can be commonly used loss functions, \emph{e.g.}, cross-entropy loss function.

To sum up, we built our method on the conventional process of the model stealing attack (step 1), and proposed a CAM-driven erasing strategy (step 2.1) and a RE-based self-KD module (step 2.2) unified by a novel erasing method. 
The former strategy digs out missing information between classes and aligns the attention while the latter module helps to mitigate overfitting and enhance the generalization. 
We name the whole framework as \textit{black-box dissector} and present the algorithm detail of it in Alg.\,\ref{alg1}.

\section{Experiments}
\subsection{Experiment settings}
In this subsection, we introduce our experiment settings, including victim model, model architectures, attack dataset and training process.

\textbf{Victim model.} The victim models we used (ResNet-34~\cite{he2016deep}) are trained on four datasets, namely,  CIFAR10~\cite{krizhevsky2009learning}, SVHN~\cite{netzer2011reading}, Caltech256~\cite{griffin2007caltech}, and CUBS200~\cite{wah2011caltech}, and their test accuracy are $91.56\%$, $96.45\%$, $78.40\%$, and $77.10\%$, respectively. 
All models are trained using the SGD optimizer with momentum (of 0.5) for 200 epochs with a base learning rate of 0.1 decayed by a factor of 0.1 every 30 epochs. 
In order to create an online deployment scenario, these models are all: image in, one-hot predictions out. 
Following~\cite{orekondy19knockoff,pal2020activethief,zhou2020dast}, we use the same architecture for the substitute model and will analyze the impact of different architectures.

\textbf{Attack dataset.} 
We use $1.2M$ images without labels from the ILSVRC-2012 challenge~\cite{russakovsky2015imagenet} as the attack dataset. 
In a real attack scenario, the attacker may use pictures collected from the Internet, and the ILSVRC-2012 dataset can simulate this scenario well. 
Note that we resize all images in the attack dataset to fit the size of the target datasets, which is similar to the existing setting \cite{orekondy19knockoff,pal2020activethief,zhou2020dast}.

\textbf{Training process.} 
We use the SGD optimizer with momentum (of 0.9) for 200 epochs and a base learning rate of $0.02\times{\frac{batchsize}{128}}$ decayed by a factor of 0.1 every 60 epochs. The weight decay is set to $5\times10^{-4}$ for small datasets (CIFAR10~\cite{krizhevsky2009learning} and SVHN~\cite{netzer2011reading}) and $0$ for others. 
We set up a query sequence $\{0.1\mathrm{K}, 0.2\mathrm{K}, 0.5\mathrm{K}, 0.8\mathrm{K}, 1\mathrm{K}, 2\mathrm{K}, 5\mathrm{K}, 10\mathrm{K}, 20\mathrm{K}, 30\mathrm{K}\}$ as the iterative maximum query budget, and stop the sampling stage whenever reaching the budget at each iteration. For fairness, all experiments will be conducted in accordance with this sequence. And, the model is trained from scratch for each iteration.

\textbf{Baselines and evaluation metric.}
We mainly compare our method with KnockoffNets~\cite{orekondy19knockoff} and ActiveThief~\cite{pal2020activethief}. For KnockoffNets, we use the source codes provided kindly by the authors. 
Follow~\cite{jagielski2020high}, we mainly report the test accuracy (Acc) as the evaluation metric. 
We also report the \emph{Agreement} metric proposed by~\cite{pal2020activethief} which counts how often the prediction of the substitute model is the same as the victim's as a supplement.


\subsection{Experiment results}

We first report the performance of our method compared with previous methods. 
Then, we analyze the performance of our method when encountering two SOTA defense methods (\emph{i.e.}, the adaptive misinformation~\cite{kariyappa2020defending} and the prediction poisoning~\cite{orekondy2019prediction}) and real-world online APIs.
After that, we conduct ablation experiments to analyze the contribution of each module. 
Finally, we also analyze the effect of different model structures and demonstrate the transferability of adversarial samples generated on substitute models obtained by different methods.

\begin{figure}[t]
    \centering
    \includegraphics[width=\textwidth]{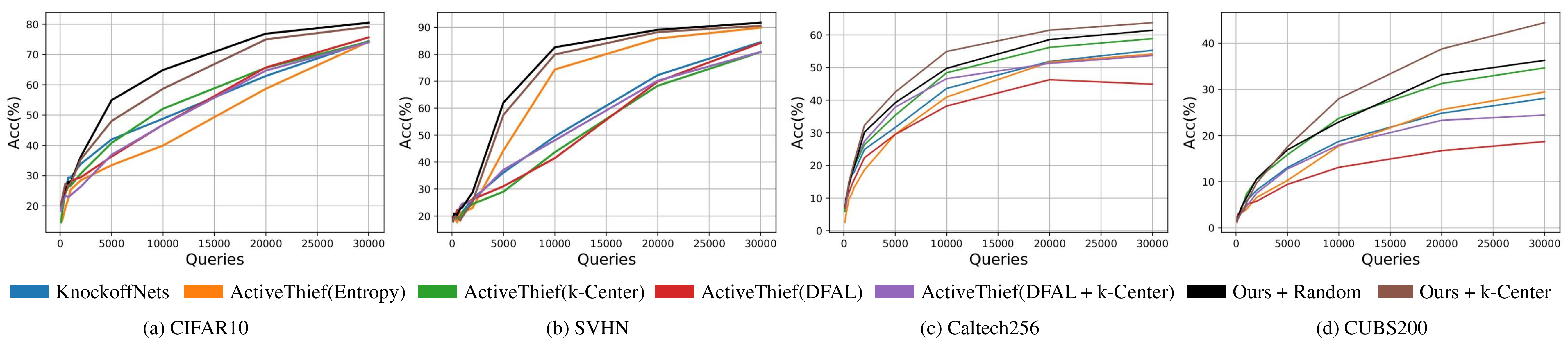}
    \caption{Curves of the test accuracy versus the number of queries.}
    \label{fig:result}
\end{figure}

\textbf{Effectiveness of our method.} As in Tab.\,\ref{tab1:result}, the test accuracy and agreement of our method are all better than the previous methods. 
We also plot the curves of the test accuracy versus the number of queries in Fig.\,\ref{fig:result}. 
The performance of our method consistently outperforms other methods throughout the process. 
Since our method does not conflict with the previous sample selection strategy, they can be used simultaneously to further improve the performance of these attacks. 
Here, we take the k-Center algorithm as an example. 
Note that, with or without the sample selection strategy, our method beats the previous methods by a large margin. 
Particularly, the test accuracies of our method are 4.85\%, 1.72\%, 3.88\%, and 8.27\% higher than the previous best method, respectively. 
And the agreement metric shares similar results. 
It is also interesting that it is less necessary to use the k-Center algorithm on datasets with a small number of classes (\emph{i.e.}, CIFAR10 and SVHN). 
While for the datasets with a large number of classes, the k-Center algorithm can make the selected samples better cover each class and improve the effectiveness of the method.

\begin{table}[t]
\centering
\caption{
The agreement and test accuracy (in \%) of each method under 30k queries. For our model, we report the average accuracy as well as the standard deviation computed over 5 runs. (\textbf{Boldface}: the best value, \textit{italics}: the second best value.)
}
\resizebox{\textwidth}{!}{
\begin{tabular}{lcccccccc}
\hline
\multirow{2}{*}{Method}    & \multicolumn{2}{c}{CIFAR10}         & \multicolumn{2}{c}{SVHN}            & \multicolumn{2}{c}{Caltech256}      & \multicolumn{2}{c}{CUBS200}         \\
\cline{2-9}
                           & Agreement        & Acc              & Agreement        & Acc              & Agreement        & Acc              & Agreement        & Acc              \\
\hline
KnockoffNets                & 75.32          & 74.44          & 85.00          & 84.50          & 57.64          & 55.28          & 30.01          & 28.03          \\
ActiveThief(Entropy)       & 75.26          & 74.21          & 90.47          & 89.85          & 56.28          & 54.14          & 32.05          & 29.43          \\
ActiveThief(k-Center)      & 75.71          & 74.24          & 81.45          & 80.79          & 61.19          & 58.84          & 37.68          & 34.64          \\
ActiveThief(DFAL)          & 76.72          & 75.62          & 84.79          & 84.17          & 46.92          & 44.91          & 20.31          & 18.69          \\
ActiveThief(DFAL+k-Center) & 74.97          & 73.98          & 81.40          & 80.86          & 55.70          & 53.69          & 26.60          & 24.42          \\
Ours+Random                       & \textbf{82.14}$\pm$0.16 & \textbf{80.47}$\pm$0.02 & \textbf{92.33}$\pm$0.47 & \textbf{91.57}$\pm$0.29 & \textit{63.61}$\pm$0.53 & \textit{61.41}$\pm$0.39 & \textit{39.07}$\pm$0.26 & \textit{36.28}$\pm$0.44 \\
Ours+k-Center              & \textit{80.84}$\pm$0.21 & \textit{79.27}$\pm$0.15 & \textit{91.47}$\pm$0.09 & \textit{90.68}$\pm$0.14 & \textbf{66.34}$\pm$0.52 & \textbf{63.75}$\pm$0.49 & \textbf{48.46}$\pm$0.55 & \textbf{44.43}$\pm$0.42 \\
\hline
\end{tabular}
}
\label{tab1:result}
\end{table}

\begin{table}[t]
\centering
\caption{Ability to evade the state-of-the-art defense methods (adaptive misinformation and prediction poisoning) on CIFAR10 dataset. 
The larger the threshold, the better the defence effect while the low victim model's accuracy (threshold 0 means no defence).
Our method evades the defense best, and the self-KD part makes a great difference.
}
\begin{tabular}{lccccccc}
\hline
Method & No defence & \multicolumn{3}{c}{Adaptive misinformation}    &     \multicolumn{2}{c}{Prediction poisoning}          \\
\hline
Threshold                        & 0                & 0.5    & 0.7    & 0.9 & 0.5 & 0.8       \\
\hline
KnockoffNets                                & 74.44\%          & 74.13\% & 73.61\% & 54.98\% & 71.83\% & 58.01\% \\
ActiveThief(k-Center)                       & 74.24\%          & 69.14\% & 59.78\% & 50.19\% & 73.75\% & 60.89\% \\
ActiveThief(Entropy)                        & 74.21\%          & 71.61\% & 64.84\% & 51.07\% & 72.07\% & 65.83\% \\
Ours                                        & \textbf{80.47\%} & \textbf{79.95\%} & \textbf{78.25\%} & \textbf{74.40\%} & \textbf{80.01\%} & \textbf{79.23\%} \\
Ours w/o self-KD                        & 79.02\%          & 78.66\% & 73.61\% & 61.81\% & 78.87\% & 76.49\% \\
\hline
victim model                                & 91.56\%          & 91.23\% & 89.10\% & 85.14\% & 91.56\% & 89.45\% \\
\hline
\end{tabular}
\label{tab:mis}
\end{table}

\textbf{Ability to evade the SOTA defense method.} 
Here we evaluate two SOTA perturbation-based defense method, the adaptive misinformation~\cite{kariyappa2020defending} and the prediction poisoning~\cite{orekondy2019prediction}. 
The adaptive misinformation~\cite{kariyappa2020defending} introduces an Out-Of-Distribution (OOD) detection module based on the maximum predicted value and punishes the OOD samples with a perturbed model $f^{\prime}(\cdot;\theta^{\prime})$.
This perturbed model ${f}^{\prime}(\cdot;{\theta}^{\prime})$ is trained with $\arg \min_{{\theta}^{\prime}} \mathbb{E}_{(x,y)}[-\log(1-{f}^{\prime}(x;{\theta}^{\prime})_y)]$ to minimize the probability of the correct class. 
%
Finally, the output will be:
\begin{equation}
    \begin{aligned}
        & y^{\prime} =(1-\alpha)f(x;\theta)+(\alpha){f}^{\prime}(x;{\theta}^{\prime}), \\
    \end{aligned}
\end{equation}
where $\alpha=1/(1+e^{\nu (\max f(x;\theta)-\tau)})$ with a hyper-parameter $\nu$ is the coefficient to control how much correct results will be returned, and $\tau$ is the threshold used for OOD detection. 
The model returns incorrect predictions for the OOD samples without having much impact on the in-distribution samples.
The prediction poisoning~\cite{orekondy2019prediction} is also a perturbation-based defense method, which perturb the posterior probabilities $y$ to make the adversarial gradient signal that maximally deviates from the original gradient. 
As shown by the following equation:
\begin{equation}
\max _{\tilde{y}}\left\|\frac{G^{T} \tilde{y}}{\left\|G^{T} \tilde{y}\right\|_{2}}-\frac{G^{T} y}{\left\|G^{T} y\right\|_{2}}\right\|_{2}^{2}
\end{equation}
where $G={\nabla}_{w}\log F(x;w)$, $y$ is the posterior probabilities and $\tilde{y}$ is the perturbed posterior probabilities.

We choose three values of the threshold $\tau$ in the adaptive misinformation and two values of the threshold $\epsilon$ in the prediction poisoning to compare the effects of our method with the previous methods. The threshold value of $0$ means no defence. The result is shown in Tab.\,\ref{tab:mis}. 
Compared with other methods, adaptive misinformation and prediction poisoning are almost ineffective  to our method. 
Furthermore, we find that if we remove the self-KD in our method, the performance is greatly reduced. 
We conclude that this is because these two defence methods add noise labels to the substitute model's training dataset, and self-KD can alleviate the overfitting of the substitute model to the training dataset, making these two defence methods not effective enough.



\begin{figure}[tb]
    \centering
    \includegraphics[width=\columnwidth]{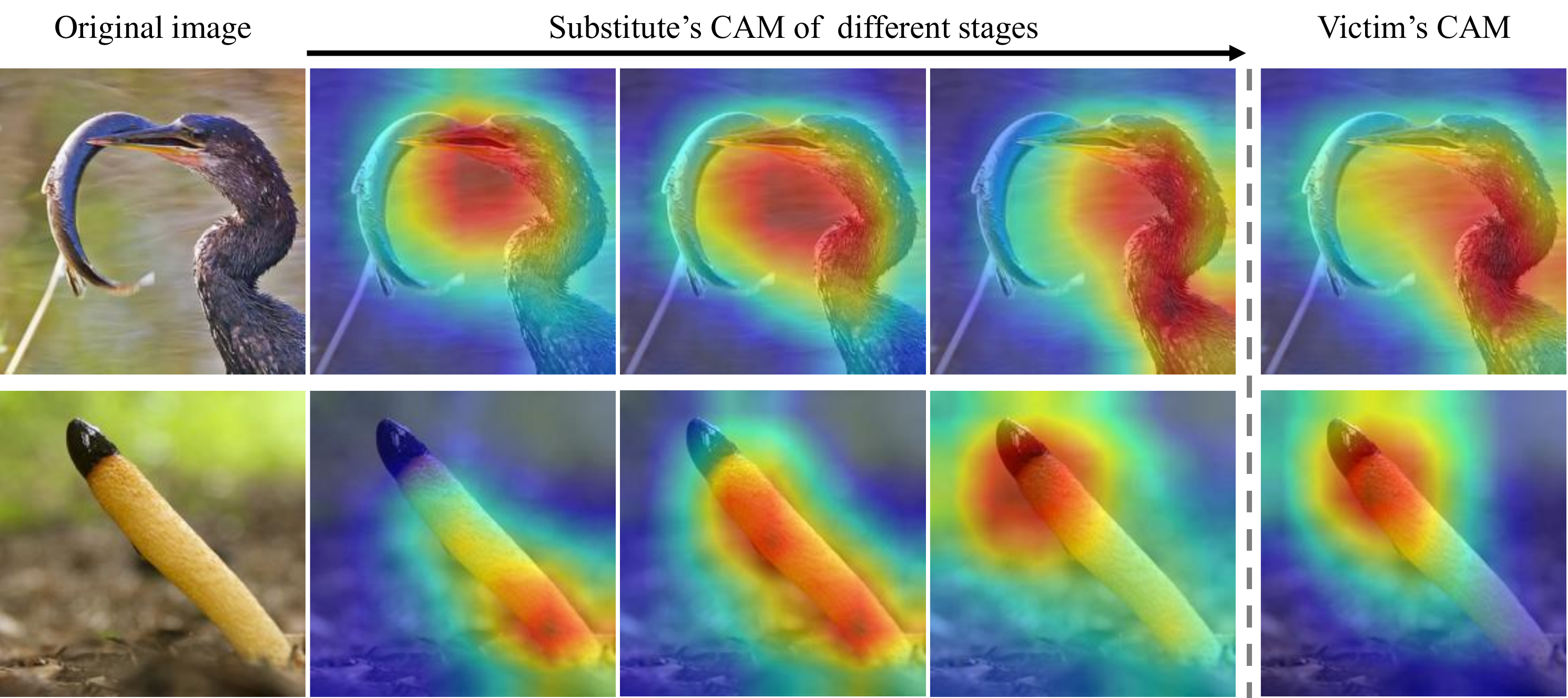}
    \caption{
    The visualized attention maps of the victim model and different stages substitute models using the Grad-CAM.
    Along with the training stages, the attention map of the substitute model tends to fit the victim model's.
    }
    \label{fig:cam}
\end{figure}

\begin{table}[t]
\centering
\caption{
An ablation experiment showing the effectiveness of the two modules we designed on CIFAR10 dataset under 30k queries. We use some commonly used regularization methods to replace the two modules we designed, and the results show that the two modules are better than the traditional regularization methods.
}
\begin{tabular}{lc}
\hline
Method & ACC   \\
\hline
Ours & 80.47\% \\
+ 2 $\times$ weight decay & 81.65\% \\
\hline
- CAM-driven erasing & 76.12\% \\
- CAM-driven erasing + CutOut & 77.11\% \\
\hline
- self-KD & 79.02\% \\
- self-KD + 2 $\times$ weight decay & 80.09\% \\
- self-KD + CutOut & 78.91\% \\
- self-KD + label smoothing (0.9) & 78.22\% \\
- self-KD + label smoothing (0.8) & 77.46\% \\
\hline
\end{tabular}
\label{tab:ablation}
\end{table}

\begin{figure}[tb]
    \centering
    \subfigure[Ablation study]{
    \includegraphics[width=0.47\columnwidth]{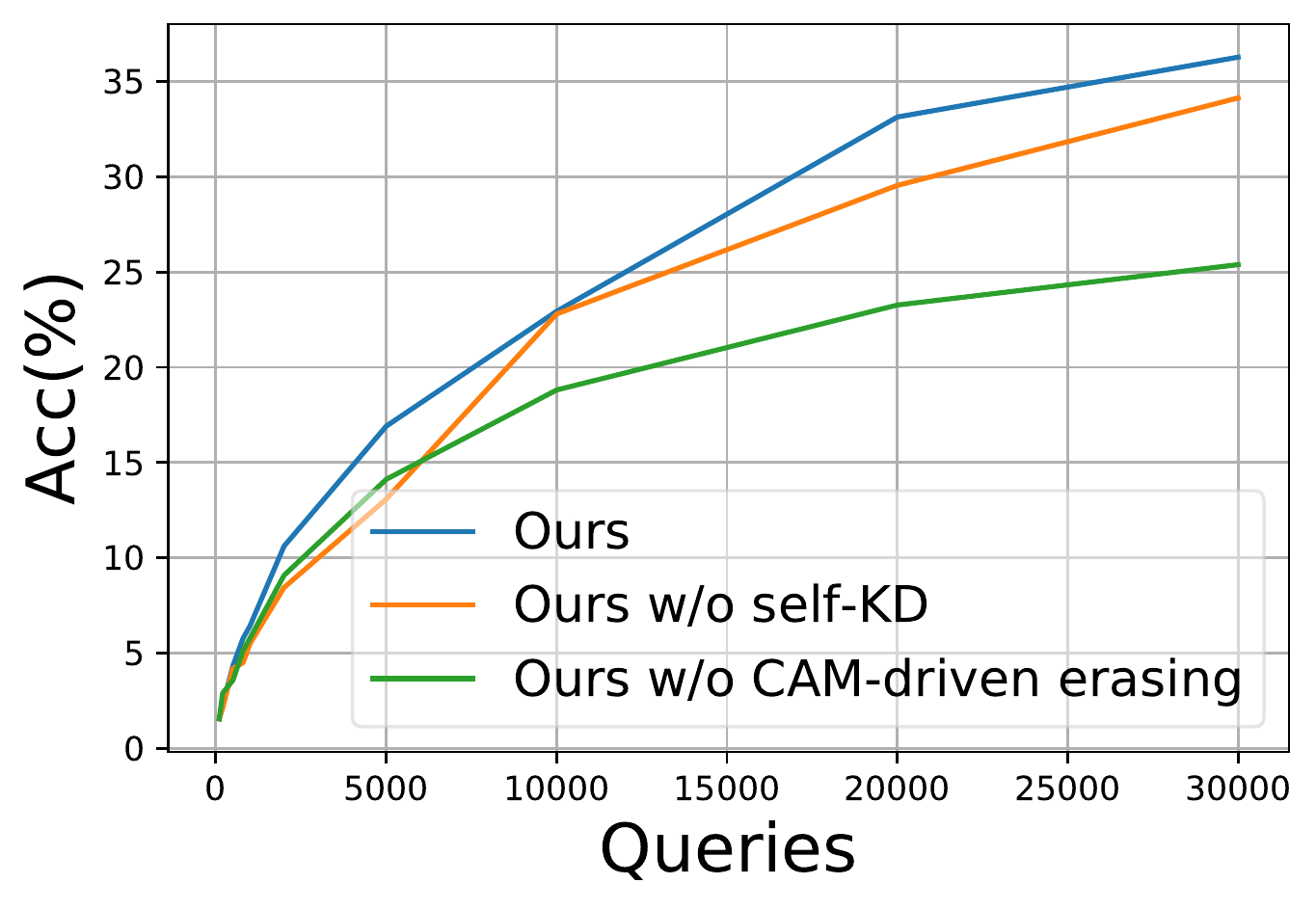}
    }
    \subfigure[AWS online API]{
    \includegraphics[width=0.47\columnwidth]{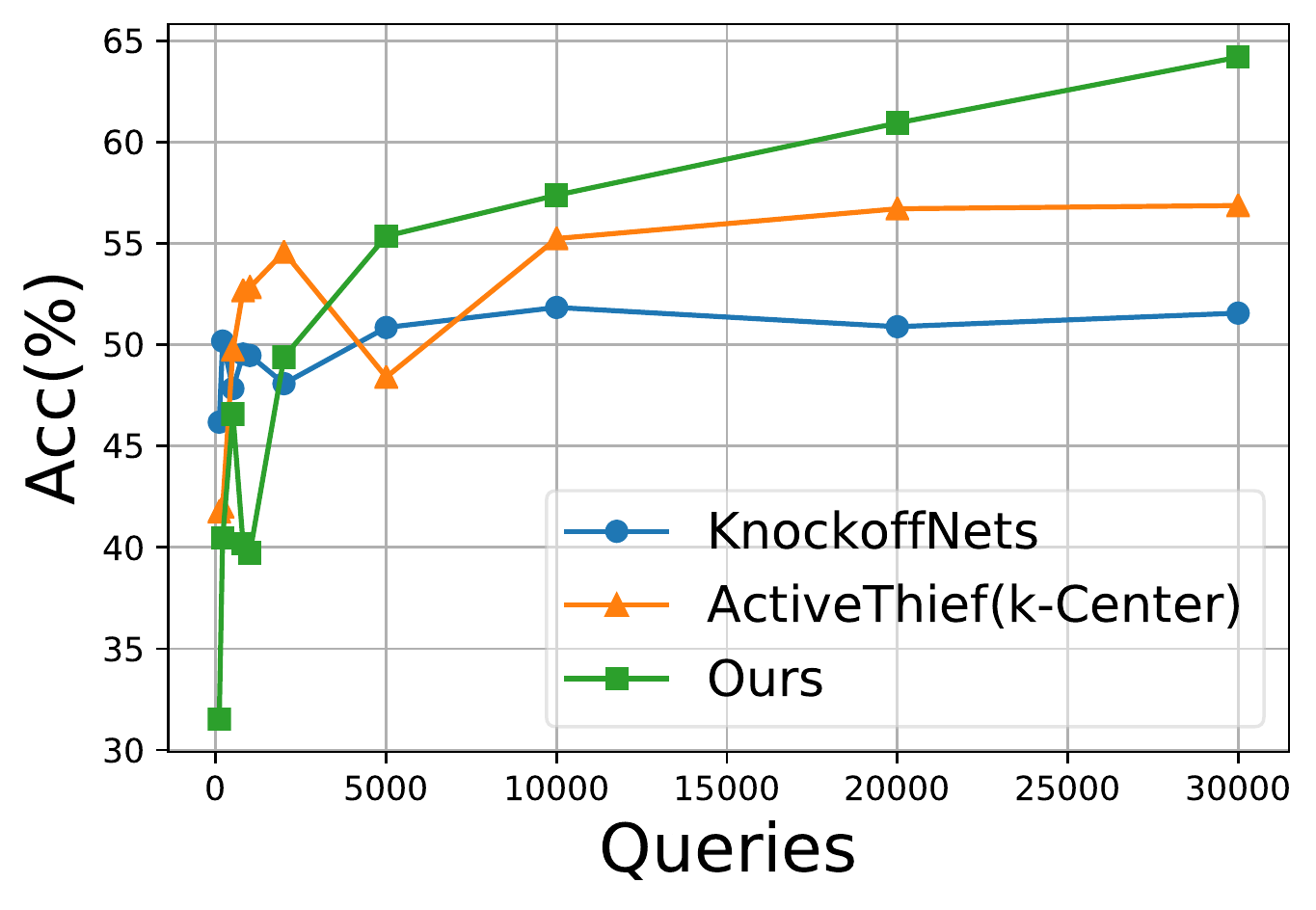}
    }
    \caption{
    (a) An ablation study on CUBS200 dataset for the contribution of the CAM-driven erasing and the self-KD. 
    (b) The experiment on a real-word online API.}
    \label{fig:ablation}
\end{figure}

\textbf{Ablation study.} 
To evaluate the contribution of different modules in our method, we conduct the ablation study on CIFAR10 dataset and show the results in Tab.\,\ref{tab:ablation}. 
We first separately remove the two modules we designed to verify their role. 
If the CAM-driven erasing strategy is removed, the performance of our method will be greatly reduced, showing that it has an indispensable position in our method. 
We also give some visual examples in Fig.\,\ref{fig:cam} to demonstrate that this strategy can help align the attention of two models. 
As depicted in the Fig.\,\ref{fig:cam}, at the beginning time, the substitute model learns the wrong attention map. 
Along with the iterative training stages, the attention area of the substitute model tends to fit the victim model's, which conforms to our intention. 
We further remove the self-KD module to evaluate its performance. 
It can be found from Fig.\,\ref{fig1} and Tab.\,\ref{tab:ablation} that the self-KD can improve the generalization of our method and further improve the performance. 
Later, in order to prove that these modules are better than some commonly used regularization methods, such as CutOut, label smoothing, we use these methods to replace the modules we designed. 
Note that the weight decay we used before followed the setting of baseline, so here we also test the effect of a large weight decay. 
The results are shown in Tab.\,\ref{tab:ablation}, where " $2 \times$ weight decay" represents the expansion of the weight decay to twice the original and the "label smoothing ($\alpha$)" means smooth the hard-label according to the hyperparameter $\alpha$. 
First, we replace the CAM-driven erasing with random erasing (CutOut), which brings 3.36\% performance degradation. We believe that using Grad-CAM as a prior is more effective than random. 
Then we use data augmentation (CutOut) and label smoothing to replace the self-KD, while both show less competitive. We conclude that they destroy the information need by the CAM-driven erasing, e.g., erasing the attention map or hiding information in other classes by making them equal. The result also shows the effectiveness of the self-KD module we designed. 
In addition, we also perform a simple ablation experiment on the CUBS200 dataset. The results are shown in Fig.\,\ref{fig:ablation} (a) and are similar to those on the CIFAR10 dataset.

\textbf{Stealing functionality of a real-world API.} 
We validate our method is applicable to real-world APIs. 
The AWS Marketplace is an online store that provides a variety of trained ML models for users. It can only be used in the form of a black-box setting. 
We choose a popular model (waste classifier\,\footnote{For the purpose of protecting privacy, we hide the specific information of the victim model.}) as the victim model. 
We use ILSVRC-2012 dataset as the attack dataset and choose another small public waste classifier dataset\,\footnote{\url{https://github.com/garythung/trashnet}}, containing $2,527$ images as the test dataset. 
The hyperparameter settings remain the same as before. 
As in Fig.\,\ref{fig:ablation} (b), the substitute model obtained by our method achieves $12.63\%$ and $7.32\%$ improvements in test accuracy compared with two previous methods, which show our method has stronger practicality in the real world.

\begin{table}[tb]
    \centering
    \caption{Transferability of adversarial samples generated with PGD attack on the substitute models.}
    \begin{tabular}{lccccc}
    \hline
    \multirow{2}{*}{Method} & \multicolumn{5}{c}{Substitute's architecture}     \\
    \cline{2-6}
                            & ResNet-34 & ResNet-18 & ResNet-50 & VGG-16 & DenseNet      \\
    \hline
    KnockoffNets            & 57.85\%    & 63.33\%    & 52.04\%    & 42.88\% & 60.77\%  \\
    ActiveThief(k-Center)   & 57.44\%    & 57.90\%    & 57.01\%    & 16.49\% & 60.72\%  \\
    ActiveThief(Entropy)    & 63.56\%    & 66.76\%    & 58.19\%    & 55.43\% & 62.05\%  \\
    Ours                    & \textbf{76.63\%}    & \textbf{74.10\%}   & \textbf{74.28\%}    & \textbf{67.03\%} & \textbf{66.96\%}  \\
    \hline
    \end{tabular}
    \label{tab:adv}
\end{table}

\textbf{Transferability of adversarial samples.}
Though with the dominant performance on a wide range of tasks, deep neural networks are shown to be vulnerable to imperceptible perturbations, \emph{i.e.}, adversarial examples~\cite{szegedy2013intriguing,fang2022learning}. 
Since the model stealing attack can obtain a functionally similar substitute model, some previous works (\emph{e.g.}, JBDA~\cite{papernot2017practical},  DaST~\cite{zhou2020dast} and ActiveThief~\cite{pal2020activethief}) used this substitute model to generate adversarial samples and then performed the transferable adversarial attack on the victim model. 
We argue that a more similar substitute model leads to a more successful adversarial attacks. 
We test the transferability of adversarial samples on the test set of the CIFAR10 dataset. 
Keeping the architecture of the victim model as the ResNet-34, we evaluate the attack success rate of adversarial samples generated from different substitute models (\emph{i.e.}, ResNet-34, ResNet-18, ResNet-50~\cite{he2016deep}, VGG-16~\cite{simonyan2014very}, DenseNet~\cite{huang2017densely}). 
All adversarial samples are generated using Projected Gradient Descent (PGD) attack~\cite{madry2017towards} with maximum $L_\infty$-norm of perturbations as $8/255$. 
As shown in Tab.\,\ref{tab:adv}, the adversarial samples generated by our substitute models have stronger transferability in all substitute's architectures, with $4.91\%-16.20\%$ improvements than other methods. 
This again proves that our method is more practical in real-world scenarios.

\section{Conclusion}
We investigated the problem of model stealing attacks under the hard-label setting and pointed out why previous methods are not effective enough. 
We presented a new method, termed \emph{black-box dissector}, which contains a CAM-driven erasing strategy and a RE-based self-KD module. 
We showed its superiority on four widely-used datasets and verified the effectiveness of our method with defense methods, real-world APIs, and the downstream adversarial attack. 
Though focusing on image data in this paper, our method is general for other tasks as long as the CAM and similar erasing method work, \emph{e.g.}, synonym saliency words replacement for NLP tasks~\cite{dong2021towards}. 
We believe our method can be easily extended to other fields and inspire future researchers. 
Model stealing attack poses a threat to the deployed machine learning models. 
We hope this work will draw attention to the protection of deployed models and furthermore shed more light on the attack mechanisms and prevention methods.
Additionally, transformer-based classifiers are becoming hot, and their security issues should also be paid attention to. 
This kind of classifier divides the images into patches and our method works by erasing parts of images, it is more convenient for us to align the attention map by masking the patch and mine the missing information. 
We will validate this idea in the further work.

\section*{Acknowledgments}
This work was supported by the National Science Fund for Distinguished Young Scholars (No.62025603), the National Natural Science Foundation of China (No. U21B2037, No. 62176222, No. 62176223, No. 62176226, No. 62072386, No. 62072387, No. 62072389, and No. 62002305), Guangdong Basic and Applied Basic Research Foundation (No.2019B1515120049), and the Natural Science Foundation of Fujian Province of China (No.2021J01002).

\clearpage
%
%
\bibliographystyle{splncs04}
\bibliography{ref}
\end{document}